\documentclass{article}

\usepackage{arxiv}

\usepackage[utf8]{inputenc} % allow utf-8 input
\usepackage[T1]{fontenc}    % use 8-bit T1 fonts
\usepackage{hyperref}       % hyperlinks
\usepackage{url}            % simple URL typesetting
\usepackage{booktabs}       % professional-quality tables
\usepackage{amsfonts}       % blackboard math symbols
\usepackage{nicefrac}       % compact symbols for 1/2, etc.
\usepackage{microtype}      % microtypography
\usepackage{lipsum}
\usepackage{graphicx}
\usepackage{adjustbox}

\usepackage{glossaries}
\setacronymstyle{long-short}
\newacronym{ML}{ML}{machine learning}
\newacronym{XAI}{XAI}{explainable artificial intelligence}
\newacronym[longplural={counterfactual explanations}]{CFE}{CFE}{counterfactual explanation}
\newacronym{AMT}{AMT}{Amazon Mechanical Turk}
\newacronym{BMBF}{BMBF}{Federal Ministry of Education and Research Germany}
\newacronym{DLR}{DLR}{German Aerospace Centre}

\usepackage{xspace}
\newcommand*{\eg}{e.g.\@\xspace}
\newcommand*{\ie}{i.e.\@\xspace}
\newcommand{\refeq}[1]{Eq.~\eqref{#1}}

\newcommand{\set}[1]{\mathcal{#1}}

\DeclareMathOperator*{\loss}{{\ell}}

\newcommand{\RNmath}{\mathbb{R}}

\DeclareMathOperator*{\regularization}{\ensuremath{{\theta}}}
\newcommand{\x}{\ensuremath{\vec{x}}}

\newcommand{\setY}{\ensuremath{\set{Y}}}

\newcommand{\xcf}{\ensuremath{{\vec{x}_{\text{cf}}}}}

\newcommand{\ycf}{\ensuremath{y'}}

\newcommand{\dimsym}{d}

\newcommand{\classifier}{\ensuremath{h}}

\usepackage[backend=biber,
            sorting=nyt,
            maxcitenames=2,
            maxbibnames=8,
            backref=true,
            natbib=true]{biblatex}
\usepackage{subcaption}
\usepackage{makecell}
\usepackage{booktabs}
\usepackage{multirow}
\usepackage{tabularx}

\title{Keep Your Friends Close and Your Counterfactuals Closer: Improved Learning From Closest Rather Than Plausible Counterfactual Explanations in an Abstract Setting}

\author{
Ulrike Kuhl \\
  \textit{CITEC} \\
  Bielefeld University\\
  Bielefeld, Germany \\
  \texttt{ukuhl@techfak.uni-bielefeld.de} \\
   \And
 Andr\'e Artelt \\
  \textit{CITEC} \\
  Bielefeld University\\
  Bielefeld, Germany \\
  \texttt{aartelt@techfak.uni-bielefeld.de} \\
  \And
 Barbara Hammer \\
  \textit{CITEC} \\
  Bielefeld University\\
  Bielefeld, Germany \\
  \texttt{bhammer@techfak.uni-bielefeld.de} \\
}

\begin{document}
\maketitle
\begin{abstract}
\Glspl{CFE} highlight what changes to a model's input would have changed its prediction in a particular way. \Glspl{CFE} have gained considerable traction as a psychologically grounded solution for \gls{XAI}. 
Recent innovations introduce the notion of computational plausibility for automatically generated \glspl{CFE}, enhancing their robustness by exclusively creating plausible explanations. %, limiting the set of possible counterfactuals to the training data.
However, practical benefits of such a constraint on user experience and behavior is yet unclear.
In this study, we evaluate objective and subjective usability of computationally plausible \glspl{CFE} in an iterative learning design targeting novice users.
We rely on a novel, game-like experimental design, revolving around an abstract scenario.
Our results show that novice users actually benefit less from receiving computationally plausible rather than closest \glspl{CFE} that produce minimal changes leading to the desired outcome.
Responses in a post-game survey reveal no differences in terms of subjective user experience between both groups.
Following the view of psychological plausibility as comparative similarity, this may be explained by the fact that users in the closest condition experience their \glspl{CFE} as more psychologically plausible than the computationally plausible counterpart.
In sum, our work highlights a little-considered divergence of definitions of computational plausibility and psychological plausibility, critically confirming the need to incorporate human behavior, preferences and mental models already at the design stages of \gls{XAI} approaches. In the interest of reproducible research, all source code, acquired user data, and evaluation scripts of the current study are available: \url{https://github.com/ukuhl/PlausibleAlienZoo}
\end{abstract}

% reset all acronyms
\glsresetall

\section{Introduction}\label{sec:introduction}

\insert\footins{
  \normalfont\footnotesize
  \interlinepenalty\interfootnotelinepenalty
  \splittopskip\footnotesep \splitmaxdepth \dp\strutbox
  \floatingpenalty10000 \hsize\columnwidth
  Permission to make digital or hard copies of part or all of this work for personal or classroom use is granted without fee provided that copies are not made or distributed for profit or commercial advantage and that copies bear this notice and the full citation on the first page. Copyrights for third-party components of this work must be honored. For all other uses, contact the owner/author(s).\\ 
  \textit{FAccT ’22, June 21–24, 2022, Seoul, Republic of Korea}\\
  \copyright~Copyright held by the owner/author(s).\\
  ACM ISBN 978-1-4503-9352-2/22/06.\\
  \url{https://doi.org/10.1145/3531146.3534630}}

% XAI is all the craze!
Explaining one's behavior to another person is a critical element in human social interaction. 
We depend on explanations to improve our understanding, ultimately building a stable mental model as basis for prediction and control~\citep{heider_psychology_1958}. 
The need to effectively explain not just human action, but also the behavior of automated systems and their underlying \gls{ML} models, has received increasing attention in recent years. This development gave rise to the increasing interest in \gls{XAI} as a research field. 
% Supposed interpretable models are there, exp CFs, but user exaluations are often neglected.
Consequently, the \gls{XAI} community has seen a veritable surge of technical accounts on how to realize explainability for \gls{ML}~\citep{guidotti_survey_2019}. 

Motivated by a seminal review by~\citeauthor{miller_explanation_2019} advocating a user-centered focus on explainability, \glspl{CFE} gained particular prominence as a supposedly useful, human-accessible solution~\citep{miller_explanation_2019,keane_if_2021}.
\glspl{CFE} provide \textit{what-if} feedback to the user, \ie, information on what changes in the input elicit a change of an automated decision (\ie, ``if you had worn a mask, you would not have gotten ill''). % Counterfactuals can be interpreted as a recommendations of actions
However, the emerging body of work on \glspl{CFE}, and explainability of \gls{ML} models more generally, shows an alarming tendency to take the quality of the suggested explanation modes at face value~\citep{doshi-velez_towards_2017, offert_i_2017}.
A recent review of counterfactual \gls{XAI} studies reveals that only one on three studies concern themselves with user-based evaluations, often with limitations concerning statistical power and reproducibility~\citep{keane_if_2021}.

% A profound research gap: comparing different approaches:
The lack of user-based evaluations affects not only assessments of \glspl{CFE} as such, but more specifically also the evaluation of different conceptualizations for this kind of explanations.
The prevailing approach in the current literature is to compare different \gls{CFE} approaches exclusively in terms of their robustness and theoretical fairness~\citep{artelt_evaluating_2021, de_oliveira_framework_2021, white_measurable_2020}, passing over the role of the user as eventual target.  
Thus, in-depth evaluations of user experiences, elucidating the usability of \gls{CFE} variants, are yet to be done.

% Our contribution! 
The current work marks a step towards closing this fundamental research gap, focusing on the concept of plausibility.
While technical descriptions of plausible \glspl{CFE} approaches exist~\citep{smyth_few_2021,schleich_geco_2021,artelt_efficient_2022}, no user study to date has directly investigated potential benefits of enforcing an additional plausibility constraint.
Thus, we perform a well-powered user study analyzing the performance of novice users when receiving closest \glspl{CFE} exclusively defined via their proximity to the decision boundary, compared to computationally plausible \glspl{CFE} as feedback in an iterative learning design~\citep{artelt_convex_2020, artelt_efficient_2022}.

\section{Counterfactual Explanations as a Psychologically Grounded Solution for XAI}\label{subsec:psychoCFs}

% we still don't know what a good explanation is
A major challenge for \gls{XAI} is the lack of a common, straight-forward and universally applicable definition of what constitutes a good explanation.
To complicate matters, the effectiveness of an approach may depend on the reason for explaining~\citep{adadi_peeking_2018}, as well as pre-existing knowledge and experiences of users at the receiving end~\citep{van_der_waa_evaluating_2021}.

% insights from psychology can help!
In search of truly human-usable explanation modes, the \gls{XAI} community recognized the need to bridge the gap between psychology and computer science in order to draw inspiration from how humans explain in their daily social interactions~\citep{miller_explanation_2019}.
A central insight from classical psychological literature is that human explanations are typically contrastive: They emphasize (explicitly or implicitly) why a specific outcome occurred instead of another~\citep{miller_explanation_2019, lipton_contrastive_1990, lombrozo_explanation_2012, hilton_knowledge-based_1986}.

% from contrastive explanations to counterfactual thinking / counterfactural reasoning / CF explanations
This contrastive nature relates to the more general human tendency to reflect upon past events by generating possible alternatives, \ie, counterfactual thinking ~\citep{roese_counterfactual_1997}.
Empirical evidence demonstrates that humans show this \textit{what-if} mentality spontaneously~\citep{goldinger_blaming_2003}, and increasingly when facing negative outcomes or unexpected results~\citep{sanna_antecedents_1996}.
In their functional theory of counterfactual thinking,~\citeauthor{roese_functional_2017} suggest a crucial role of counterfactual thoughts to guide to formation of future intentions, thus regulating subsequent behavior~\citep{roese_functional_2017,epstude_functional_2008}.
This evidence is the root for the common supposition in \gls{XAI} that explanations formulated as counterfactuals are naturally intuitive, easy to understand, and helpful for users, often discounting the need for user evaluations~\citep{stepin_paving_2019, dandl_multi-objective_2020, guidotti_local_2018, artelt_efficient_2022}.

% how do humans generate CFEs? 
Decades of philosophical and psychological research has concerned itself with the question of how humans generate counterfactuals. 
% --> possible worlds account by Lewis
Lewis' seminal work on the topic builds on a theory of possible worlds, postulating that counterfactual statements trigger a comparison between the actual circumstances and a conceivable world in which the counterfactual statement occurred~\citep{lewis_counterfactuals_1973}.
% how do humans generate CFEs? --> mental models view, Byrne
Embedding this view into a cognitive framework of counterfactual thought, the mental models theory emphasizes the human ability to entertain two parallel representations of reality: The factual conditional, corresponding to the true state of the world, and the concurrent non-factual possibility, temporarily assumed to be true~\citep{byrne_mental_2002,byrne_precis_2007,johnson-laird_conditionals_2002,walsh_mental_2005}.
% Neuroscience evidence:
Insights from neuroimaging support this notion, demonstrating that counterfactual thinking extends mere hypothetical deliberation by recruiting additional representational processes in the brain~\citep{kulakova_processing_2013}.

% Authors often note that humans prefer 
When humans generate counterfactuals, they show remarkable regularities in terms of which aspects of the past they reconstruct. 
Humans tend to modify events that are recent~\citep{miller_temporal_1990, byrne_temporality_2000}, exceptional, while also regarding the optimal counterfactual outcome~\citep{kahneman_simulation_1982, dixon_if_2011}, and 
controllable events when undoing of fictitious outcomes~\citep{girotto_event_1991}.
Further, authors like to note that humans produce plausible rather than implausible counterfactuals~\citep{byrne_counterfactual_2016, de_brigard_coming_2013}.

% What is plausible from a psycho point of view? Hard to say
However, despite being a commonly-used notion in psychology, plausibility is difficult to define precisely.
Variable interpretations of what constitutes a plausible counterfactual exist, referring to different partially overlapping concepts.
\citeauthor{kahneman_simulation_1982} refer to hypothetical events as plausible if they are easy to imagine~\citep{kahneman_simulation_1982}. 
% plausibility as closeness / similarity / comparative similarity:
\citeauthor{lewis_counterfactuals_1973} supposes that plausible counterfactuals come from worlds that are minimally different from reality~\citep{lewis_counterfactuals_1973}. 
Building up on this idea of comparative similarity, empirical research shows perceived plausibility of a counterfactual event to be proportional to the perceived similarity between said counterfactual and the factual state~\citep{stanley_counterfactual_2017, de_brigard_perceived_2021}.

% plausibility as probability:
In addition to such a similarity-based definition, plausibility is often used synonymously with concepts of likeliness or probability~\citep{pezdek_is_2006, de_brigard_remembering_2013}. 
\citeauthor{de_brigard_remembering_2013} demonstrate that manipulations of counterfactual plausibility in terms of their likeliness changes their neural representation~\citep{de_brigard_remembering_2013}.
Their findings may indicate greater affective evaluation for counterfactuals that carry greater subjective likelihood, and thus, plausibility.
In their plausibility analysis model,~\citeauthor{connell_model_2006} expand on the idea of plausibility as probability and highlight the pivotal role of pre-existing domain knowledge, postulating that a scenario may only be plausible if it fits well to prior knowledge~\citep{connell_model_2006}.

% Concluding, and bring it together with automatic CFEs
Thus, while it is difficult to pinpoint exactly what makes a counterfactual psychologically plausible, we may recognize pivotal roles of concepts like comparative similarity and probability.
Following the user-centered focus on explainability proposed by~\citeauthor{miller_explanation_2019}~\citep{miller_explanation_2019}, incorporating these concepts would be an important step towards automatic generation of plausible, and thus more human-friendly and usable, \gls{CFE}.

\section{Computation of Closest CFEs and Plausible CFEs}\label{sec:EffCompCFs}

\citeauthor{wachter_counterfactual_2017} introduce a \gls{CFE} $\xcf\in\RNmath^\dimsym$ of an \gls{ML} model $\classifier:\RNmath^\dimsym\to\setY$ as an optimization problem~\citep{wachter_counterfactual_2017} :
\begin{equation}\label{eq:counterfactual_opt_original}
\underset{\xcf \,\in\, \RNmath^\dimsym}{\arg\min}\; \loss\big(\classifier(\xcf), \ycf\big) + C \cdot \regularization(\xcf, \x)
\end{equation}
where $\x\in\RNmath^\dimsym$ denotes the original input, the regularization $\regularization(\cdot)$ penalizes deviations from the original input $\x$ (weighted by a regularization strength $C>0$), $\ycf\in\setY$ denotes the requested output/behavior of the model $\classifier(\cdot)$ under the counterfactual $\xcf$, and $\loss(\cdot)$ denotes a loss function penalizing deviations from the requested prediction.

Thus, computing \glspl{CFE} translates to finding minimal perturbations to a model's input that alter the final prediction to a desired outcome. Given the regularization term $\regularization(\cdot)$, generated \glspl{CFE} based on this definition remain as close to the original input $\x$ as possible. 
Thus, we will refer to them as \textit{closest \glspl{CFE}} for the remainder of this work.

As one of the first approaches to model \glspl{CFE} for classical \gls{ML}, Equation~\ref{eq:counterfactual_opt_original} is the forerunner of more powerful, model specific variations, as well as many methods for solving these optimization problems~\citep{verma_counterfactual_2020,artelt_computation_2019,karimi_survey_2020}.
However, it is important to note that \textit{closest \glspl{CFE}} do not necessarily yield plausible or even realistic counterfactuals.
As a matter of fact, \textit{closest \glspl{CFE}} may look like adversarials, introducing slight changes in the input that go unnoticed by a human observer despite altering the model's output~\citep{papernot_practical_2017}. 
Whether a computed \textit{closest \glspl{CFE}} corresponds to such an adversarial depends on the model, loss function and regularization, diminishing their suitability as explanation technique~\citep{laugel_issues_2019}.
Expanding the original definition in \refeq{eq:counterfactual_opt_original} by an additional plausibility constraint circumvents these issues:
\begin{equation}
\begin{split}
\underset{\xcf \,\in\, \RNmath^\dimsym}{\arg\min}\; \loss\big(\classifier(\xcf), \ycf\big) + C \cdot \regularization(\xcf, \x) \quad\quad \text{s.t. } \xcf \in \set{P}
\end{split}
\end{equation}
where $\set{P}$ denotes the set of all plausible \glspl{CFE}.

Similar to the modeling of \textit{closest \glspl{CFE}} in Equation~\ref{eq:counterfactual_opt_original}, different realizations of these \textit{computationally plausible \glspl{CFE}} have been proposed~\citep{looveren_interpretable_2019,poyiadzi_face_2019,artelt_convex_2020}. One particular instance are density based approaches~\citep{artelt_convex_2020} that restrict a counterfactual $\xcf$ to regions of high density (\eg, estimated from the training data).
In the current work, we follow an alternative approach when providing \textit{computationally plausible \glspl{CFE}} and limit the set of possible counterfactuals to the training data as a representative set of feasible examples~\citep{poyiadzi_face_2019}.

\section{Do Novice Users Profit from Computational Plausibility in an Abstract Domain?}\label{sec:hypotheses}

The guiding question of the current work is whether \textit{computationally plausible \glspl{CFE}} have an advantage over \textit{closest \glspl{CFE}} in helping users to learn from an \gls{ML} model.
To assess this question, we rely on an interactive iterative learning task, where users repeatedly choose input values for an \gls{ML} model.
In a separate study, we successfully demonstrate the added benefit of providing closest \glspl{CFE} compared to no explanation given this experimental framework~\citep{kuhl_lets_2022}.

In the current work, users receive either \textit{computationally plausible} or \textit{closest \glspl{CFE}}, highlighting how changes in the user's previous input would have lead to better results.
The main advantage of this approach is that the interplay between repeated user action and corrective feedback enables us to assess user understanding at each stage of the process objectively through task performance.

We find it conceivable that implementing a plausibility constraint indeed improves user performance. 
Specifically, we assume that repeated exposure to items representative of the training set enables humans to build a more accurate mental model of the underlying data distribution.
To obtain general insights about the usability of different types of \glspl{CFE} as such, we recruited novice users and designed the task around an abstract scenario.
This approach has the additional advantage to mitigate any difference in domain knowledge and possible misconceptions about the task setting, potentially confounding task performance~\citep{van_der_waa_evaluating_2021}.
Thus, we formulated the following three hypotheses.

%\paragraph{Hypothesis 1} 
\textit{Hypothesis 1.} We expect \textit{computationally plausible \glspl{CFE}} to be more helpful to users tasked to discover unknown relationships in data than \textit{closest} ones, both objectively and subjectively. 
Specifically, we anticipate that participants in the plausible condition a) show greater learning success, b) become more automatic and thus quicker in the task, and c) are able to explicitly identify relevant and irrelevant input features.

%\paragraph{Hypothesis 2}
\textit{Hypothesis 2.} We expect a group difference in terms of subjective understanding.
We predict that users will differ in how far they find \glspl{CFE} useful, and in how far they can utilize them, with an advantage of \textit{computationally plausible \glspl{CFE}}. %(survey items 5, 6).
Furthermore, we posit that users imagine \textit{computationally plausible \glspl{CFE}} to be more helpful for other users. %(survey item 9).

%\paragraph{Hypothesis 3} 
\textit{Hypothesis 3.} We evaluate users' understanding of the explanations themselves, their need for support to understand, and their evaluation of timing and efficacy of \gls{CFE} presentation. As structure and presentation mode of \glspl{CFE} is kept constant across conditions, we expect not to find any differences.
This analysis tests the comparability of conditions, a key feature in any experimental user design.

Finally, we do not formulate a prediction whether groups will differ in uncovering inconsistencies in the explanations presented. This will be investigated in an additional exploratory analysis.

\section{Experimental Design}\label{sec:experimental-design}

\begin{figure}
   \centering
   \includegraphics[width=\textwidth]{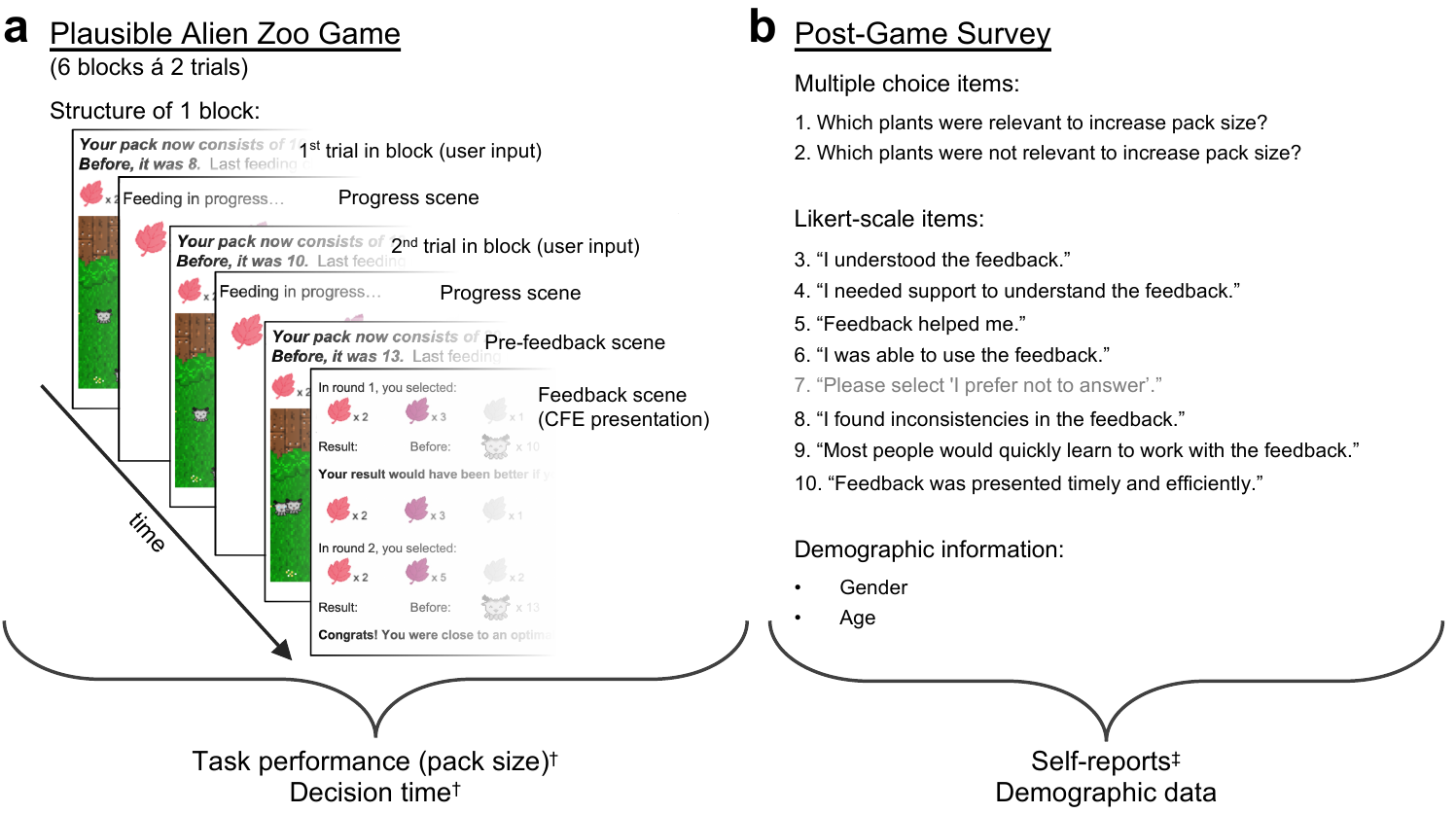}
   \caption{General overview of study procedure. (\textbf{a}) The Alien Zoo game is an iterative design with blocks containing two trials calling for user input, and finishing with a feedback scene that provides either \textit{closest} or \textit{computational plausible \glspl{CFE}} computed on user's previous input. (\textbf{b}) After 6 blocks, users enter a post-game survey collecting self-report information from participants. Likert-scale items adapted from~\citep{holzinger_measuring_2020}. Catch item marked in light gray. The lower part of both subfigures shows measures evaluates from respective parts; $^\dag$ objective measure, $^\ddag$ subjective measure.}
   \label{fig:StudyStructure}
 \end{figure}

To assess \textit{Hypotheses 1--3}, we use a novel iterative learning design revolving around an abstract scenario. Figure \ref{fig:StudyStructure} conveys the overall two-part structure of the study.

\subsection{The Alien Zoo Scenario}

We developed a game-like experimental design, the Alien Zoo~\citep{kuhl_lets_2022}.
It relies on a web-based interface to provide global access for users from diverse backgrounds, facilitating large-scale participant recruitment.
For a detailed account of the framework's conception and the rationale behind the corresponding design choices, see~\citet{kuhl_lets_2022}.

In the Alien Zoo scenario, participants imagine themselves as zookeepers for aliens. 
To feed to the aliens, participants may choose from different plants. 
However, it is not clear what plants make up a nutritious diet. 
Thus, participants need to find how to best feed the aliens. 
Participants go through several feeding cycles, choosing a combination of plants. 
After each cycle, the pack of aliens either decreases (given a bad combination of plants) or increases (given a good combination). 
In regular intervals, participants receive a summary of their past choices, together with feedback on what choice would have led to a better result (\ie a \gls{CFE}).

Assessing performance of real users in an abstract task setting, this use case corresponds to a human grounded evaluation~\citep{doshi-velez_towards_2017}.
Further, our setting falls under the ``explaining to discover'' category for explainability defined by~\citeauthor{adadi_peeking_2018}, investigating whether providing \glspl{CFE} to novice users improves their understanding of relationships in a yet unknown dataset~\citep{adadi_peeking_2018}.

\subsection{Post-Game Survey}

A post-game survey collects self-report information from participants.
Besides explicitly asking participants to point out which plants were relevant and irrelevant for the task, we use an adapted version of the System Usability Scale~\citep{holzinger_measuring_2020}, designed to measure the quality of explanations elicited by an explainable \gls{ML} system.
Participants answer a series of Likert-scale items, assessing how users feel about using our system with a focus on perceived understandibility and usability of \glspl{CFE}.
The survey closes with asking for participants' gender and age as potential confounding variables.
Figure \ref{fig:StudyStructure}b gives a complete overview of all items in the survey part, in the order participants encounter them.

\subsection{Constructs and Measurements}

We measure understanding and usability of explanations in terms of two objective behavioral variables and several subjective self-reports (Figure \ref{fig:StudyStructure} bottom). 

Regarding task performance, we assess the development of pack size in the Alien Zoo game over trials. 
This value indicates the extent of user's understanding of relevant and irrelevant features in the underlying data set, as a solid understanding leads to better feeding choices. 

Second, we measure time needed to reach a feeding decision over trials (henceforth referred to as decision time). As we assume participants to become more automatic in making their plant choice, we expect this practice effect to be reflected as decreased decision time~\citep{logan_shapes_1992}. 

We acquire self-reports via the post-game survey, assessing different aspects of participant's system understanding. 
The first two survey items ask users to identify plants they think are relevant and irrelevant for task success. 
Replies from these items allow us to measure to which extent users in different groups formed explicit knowledge of the underlying data structure. 
Further, users indicate in how far they find the explanations useful, to which degree they can make use of them, and in how far they imagine the presented \glspl{CFE} to be helpful for other users, too. These items assess user's subjective understanding.

Finally, three self-report measures check for potential confounds. 
These are items that ask users to indicate their understanding of the explanations as such, whether they feel the need for support for understanding, and their evaluation of timing and efficacy of \gls{CFE} presentation.
Given that structure and presentation mode of \glspl{CFE} is kept constant for both groups, differences would uncover unexpected variation in terms of answer style, a potential confounding variable.

\subsection{Implementation, ML Model and Data Set}
The back end of the system is written in Python3, using the sklearn package~\citep{pedregosa_scikit-learn_2011} for the \gls{ML} part. The front end employs the JavaScript-based Phaser 3, an HTML5 game framework\footnote{\url{https://phaser.io/}}.
We use a decision tree regression model for predicting the growth rate given the plants selected by the user.
Decision trees approximate the data distribution with a collection of if-then-else rules, consecutively splitting the data~\citep{shalev-shwartz_understanding_2014}.
We choose a decision tree because computing counterfactuals for this model is fairly simple~\citep{artelt_computation_2019}.
Yet, it is powerful enough to model our synthetic data set sufficiently well.
The current implementation uses the Gini splitting rule of CART~\citep{breiman_classification_1984}, with a maximum tree depth of $4$.
The decision tree corresponding to the ground truth model is build once in the beginning and remains the same for all users during the study. 

We use the code provided by the CEML package~\citep{artelt_ceml_2019} for computing \glspl{CFE}.\footnote{\url{https://github.com/andreArtelt/ceml}}
In the interest of reproducible research, all source code, acquired user data, and evaluation scripts of the current study are available.\footnote{\url{https://github.com/ukuhl/PlausibleAlienZoo}} 
% taken out to ensure anonymity
The underlying data set used for tree building consists of 5 integer features (\ie, the plants used for feeding) and 1 continuous output variable (\ie, the growth rate used as factor for computing the new pack size).
We generated this data according to the following scheme: The growth rate scales linearly with plant 2, iff plant 4 has a value of 1 or 2 OR plant 5 is not smaller than 4. 
Growth rate may take a value between 0 and 2, used as a factor for pack size in the previous round to compute the new pack size.
The initial full data set contains all possible plant -- growth rate combinations $100$ times, yielding 3 276 800 data points. 
For final model training, we sample a subset of 10 922 data points from this full set to introduce sparsity, thus ensuring that computed \textit{closest} and \textit{computationally plausible \gls{CFE}} diverge.
Note that our implementation prevents pack size from shrinking below 2.

\subsection{Participants}

The study ran in early November 2021 on \gls{AMT}.
%After performing three pilots with 10 users each to refine the experimental design, we recruited a total of 100 
After piloting, we recruited a total of 100 participants for final assessment, following an a priori sample size estimation~\citep{kumle_estimating_2021}.% Batch info neccesary? % report that users were excluded if the done it before?
~A first data quality check revealed corrupted data for four participants due to logging issues. 
Thus, we acquired four additional data sets. 
All participants gave informed electronic consent by providing clickwrap agreement prior to participation.
All participants received a reward of US\$ 4 for participation. 
The ten best performing users received an additional bonus of US\$ 2. 
Game instructions informed participants about the possibility of a bonus to motivate compliance with the experimental task~\citep{bansal_updates_2019}.
The study was approved by the Ethics Committee of Bielefeld University, Germany.

\subsection{Experimental Procedure}\label{subsec:experimental-procedure}

After accepting the task on \gls{AMT}, participants are forwarded to our web server hosting the alien zoo game.
They first encounter a page informing them about purpose, procedure and expected duration of the study, their right to withdraw, confidentiality and contact details of the primary investigator.
Users may decline to participate by closing this window.
Otherwise, they indicate their agreement via button press, opening a new page.
Unbeknownst to the user, they are randomly assigned to either the \textit{closest} or the \textit{plausible} condition when they indicate agreement.

The succeeding page provides detailed instructions to the game. 
Specifically, it shows images of the aliens, as well as the selection of plants they may choose to feed from.
Written instructions detail that it is possible to choose up to six leaves per plant in whatever combination seems desirable, and that choosing healthy or unhealthy combinations leads to increases or decreases in pack size, respectively.
Further instructions emphasize the user's task to maximize the number of aliens, so-called shubs, with the best players qualifying for a monetary bonus.
Participants are also informed that they will receive feedback on what choice would have led to a better result after two rounds of feeding.
Users begin the game by clicking a ``Start'' button, appearing with a delay of 20s at the end of the page. 

\begin{figure}[t]
   \centering
   \includegraphics[width=\textwidth]{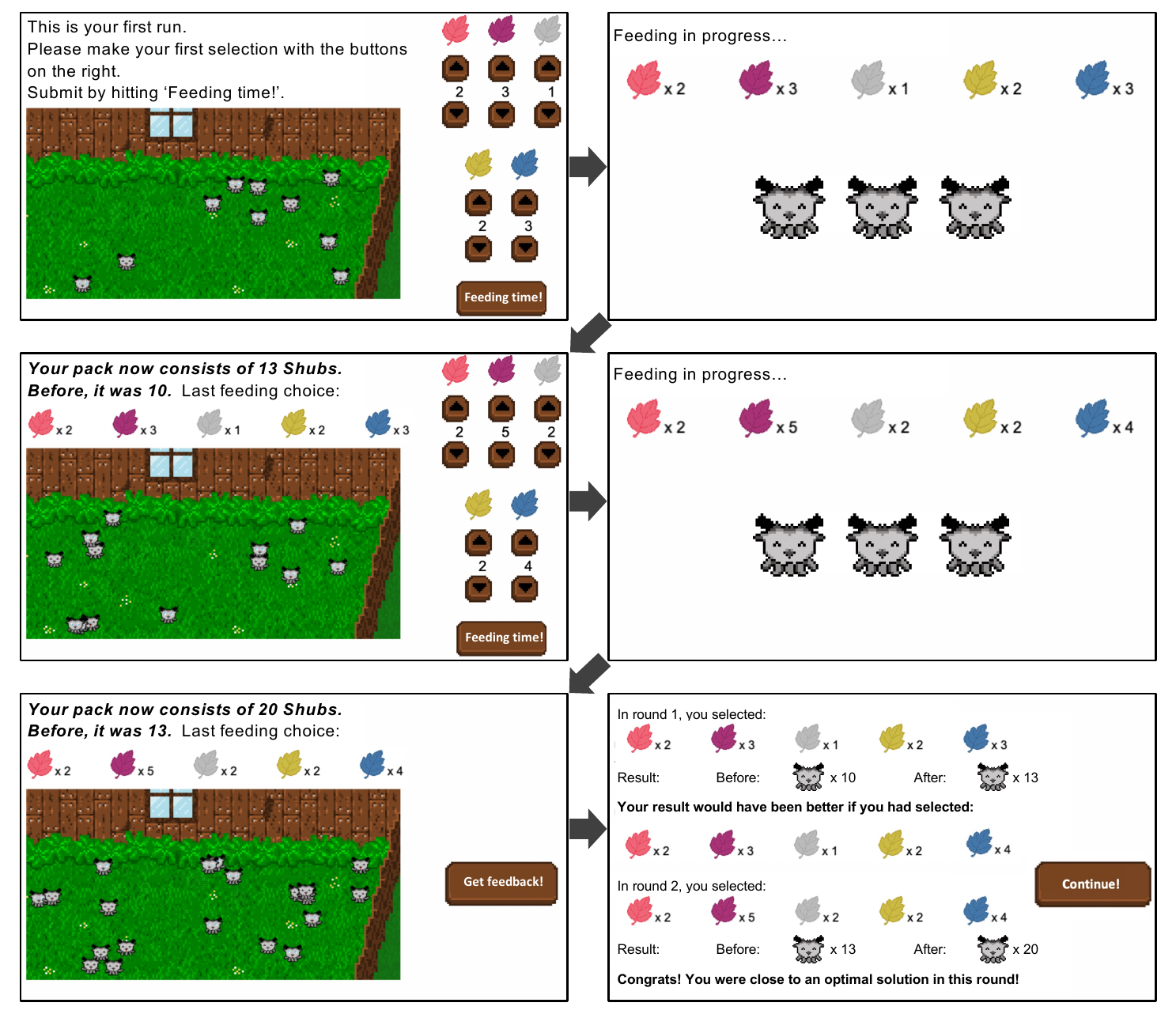}
   \caption{Exemplary user journey through the first block of the Alien Zoo game. Bold arrows indicate temporal succession of respective scenes. The figure highlights the iterative nature of the game with repeated user input and end-of-block presentation of \glspl{CFE}. Note that plant counters are set to 0 at the beginning of each padlock scene. The figure displays the state after the exemplary user inserted their current choice. For this manuscript, font size in images of scenes was increased to improve visibility.}
   \label{fig:StudyStructure2}
 \end{figure}

Upon hitting ``Start'', participants encounter a padlock scene where they can make their feeding choice (Figure \ref{fig:StudyStructure2}, top left image).
The right side of the screen displays leaves from all plant types next to upward/downward arrow buttons. 
In the first feeding round, the top of the page shows written information that clicking on the upward arrows increases the number of leaves per plant, while clicking the downward arrows has the reverse effect.
In each succeeding feeding round, the top of the page shows the current pack size, the pack size in the previous round, and the choice made in the previous round.
The page additionally shows a padlock with the current number of animated shubs.
Each participant starts of with a pack of 10 aliens.
After making their choice, participants continue by clicking a button stating ``Feeding time!'' in the bottom right corner of the screen.

Upon committing their choice, a progress scene displaying the current choice of plants and three animated aliens is shown. 
Meanwhile, the underlying \gls{ML} model uses the user input to generate the new growth rate and pack size, together with either a \textit{closest} or a \textit{computationally plausible \gls{CFE}}.
After 3s, the padlock scene appears again to show the results of their last choice. 
Following odd trials, the user may make a new selection. 
After even trials, a single ``Get feedback!'' button replaces the choice panel on the right-hand side of the screen.
Hitting the feedback button forwards a user to an overview scene displaying the feeding choices in the last two runs, the resulting changes in pack size and the counterfactuals that indicate what choices would have led to better results. 
When users made a choice that led to maximal increase in pack size such that no counterfactual could be computed, they are told that they were close to an optimal solution in that round. 
Users may move on to the next round by hitting a ``Continue!'' button appearing after 10s on the right-hand side of the screen. This delay forces users to spend some time with the information to study it. Upon continuing, users make their new choice in a new padlock scene.

The study runs over 12 feeding rounds (trials) with feedback interspersed after each second trial. 
To ensure attentiveness of users during the game, we included two additional attention trials.
After feeding rounds 3 and 7, users face a new page requesting to type in the current number of aliens in their respective packs.
Immediate feedback informs participants whether their entry was correct or not, and reminds users to pay close attention to all aspects of the game at any given time.
Subsequently, the next progress scene appears and the game continues. 

The game part of the study is complete after 12 trials.
The experimental procedure concludes with a survey assessing user's explicit knowledge on what plants were and were not relevant for improvement (items 1 and 2), as well as an adapted version of the System Causability Scale \cite{holzinger_measuring_2020} evaluating the subjective quality of explanations.
The study closes with two items assessing demographic information on gender and age.
The final page thanks users for their participation and provides a unique code to insert in \gls{AMT} to prove that they completed the study and qualify for payment. 
Further, participants may choose to visit a debriefing page with full information on study objectives and goals. 

On average, participants needed 13m:43s ($\pm$ 00m:23s SEM) from accepting the HIT on AMT to inserting their unique payment code.

\subsection{Statistical Analysis, Sample Size Calculation and Data Quality Measures}

We perform all statistical analyses using R-4.1.1~\citep{r_core_team_r_2021}, using \gls{CFE} variant (\textit{closest} or \textit{computationally plausible}) as independent variable.
Changes in performance over 12 trials measure learning rate per group (lme4 v.4\textunderscore 1.1-27.1)~\citep{bates_fitting_2015}.
In the model testing for differences in terms of user performance, the dependent variable is number of aliens generated. 
In the assessment of user's reaction time, we use trialwise decision time as dependent variable.
The final models include fixed effects of group, trial number and their interaction. The random-effect structure includes a by-subjects random intercept. 
Such linear mixed effects models account for correlations of data drawn from the same participant and missing data~\citep{detry_analyzing_2016,muth_alternative_2016}.
The analysis of variance function of the stats package in base R serves to compare model fits.
$\eta_{\text{p}}^{2}$ values denote effect sizes (effectsize v.0.5)~\citep{ben-shachar_effectsize_2020}.
Computations of pairwise estimated marginal means follow up significant main effects or interactions, with respective effect sizes reported in terms of Cohen's \textit{d}.
All post-hoc analyses reported are Bonferroni corrected to account for multiple comparisons.

We evaluate data gathered from the post-game survey depending on question type.
For the first two items assessing user's explicit knowledge of plant relevance, we test data for normality of distributions using the Shapiro-Wilk test, followed up by the non-parametric Wilcoxon-Mann-Whitney \textit{U} test in case of non-normality, and the Welch two-sample t-test otherwise for group comparisons. 
We follow the same approach to compare age and gender distributions.
We also compare user's explicit knowledge of plant relevance to the expected value given random response patterns using the non-parametric one-sample Wilcoxon signed rank test for each group separately, and report Bonferroni corrected results.
To analyze group differences of ordinal data from the Likert-style items, we rely on the non-parametric Wilcoxon-Mann-Whitney \textit{U} test.
We report effect sizes for all survey data comparisons as \textit{r}.

As a web-based study, we run the risk that some participants attempt to game the system to collect the reward without providing proper answers. 
Thus, we implement a number of data quality checks that were planned a priori.
We identify speeders based on the decision time, flagging users that spent less than 2s in the padlock scene in 4 or more trials.
We flag participants that fail to respond with the correct number of aliens in both attention trials during the game.
Furthermore, we included a catch item in the survey (\ref{fig:StudyStructure}b, item 7), flagging inattentive users.
Finally, we identify straight-lining participants who keep choosing the same plant combination despite not improving in at least three blocks, or answer with only positive or negative valence in the survey.
To uphold a high threshold for data quality, we follow a conservative approach of excluding participants that were flagged for at least one of these reasons.

\section{Results}\label{sec:results}

From the initial 100 participants, we exclude data from participants who qualified as speeders (\textit{n} = 2), failed both attention trials during the game (\textit{n} = 5), gave an incorrect response for the catch item in the survey (\textit{n} = 3), or straight-lined during the game (\textit{n} = 4) or in the survey (\textit{n} = 12), leaving data from 74 participants for final analysis (Table~\ref{tab:participants}).

\begin{table}
  \caption{Demographic information of participants.}
  \label{tab:participants}
  \resizebox{\linewidth}{!}{%
\begin{tabularx}{\textwidth}{llllllllll} 
\toprule
    & \multicolumn{4}{l}{Before quality assurance measures (\textit{N} = 100)} && \multicolumn{4}{l}{After quality assurance measures (\textit{N} = 74)} \\
\cline{2-5}\cline{7-10}
    & \textit{closest} & \textit{plausible} & \textit{U} value$^a$ & \textit{p} value && \textit{closest} & \textit{plausible} & \textit{U} value$^a$ & \textit{p} value\\ 
\hline
\textit{N}   &  50 & 50 & .. & .. && 40 & 34 & .. & .. \\
%Gender$^b$ & 17f/33m & 22f/26m/1nb/1na & 1108 & .339 && 13f/27m & 18f/15m/1nb & 554.4 & .116 \\
\multirow{2}{*}{Gender$^b$}& 17f/33m & 22f/26m/ & 1108 & .339 && 13f/27m & 18f/15m/ & 554.4 & .116\\
    & & 1nb/1na & & & & & 1nb & & \\
Age (\textit{Mdn})$^c$ & 25--34y & 25--34y & 1234 & .950 && 25--34y & 35--44y & 712.5 & .718 \\
\bottomrule
\multicolumn{9}{l}{$^a$ non-parametric Wilcoxon-Mann-Whitney \textit{U} test}\\
\multicolumn{9}{l}{$^b$ f = female, m = male, nb = non-binary / gender non-conforming, na = no gender information disclosed}\\
\multicolumn{9}{l}{$^c$ \textit{Mdn} = median age band (options: 18-24y, 25-34y, 25-34y, 35-44y, 45-54y, 55-64y, 65y and over)}
\end{tabularx}}
\end{table}

\subsection{Do Computationally Plausible CFEs Facilitate Learning?}

\begin{figure}[t]
   \centering
   \includegraphics[width=\textwidth]{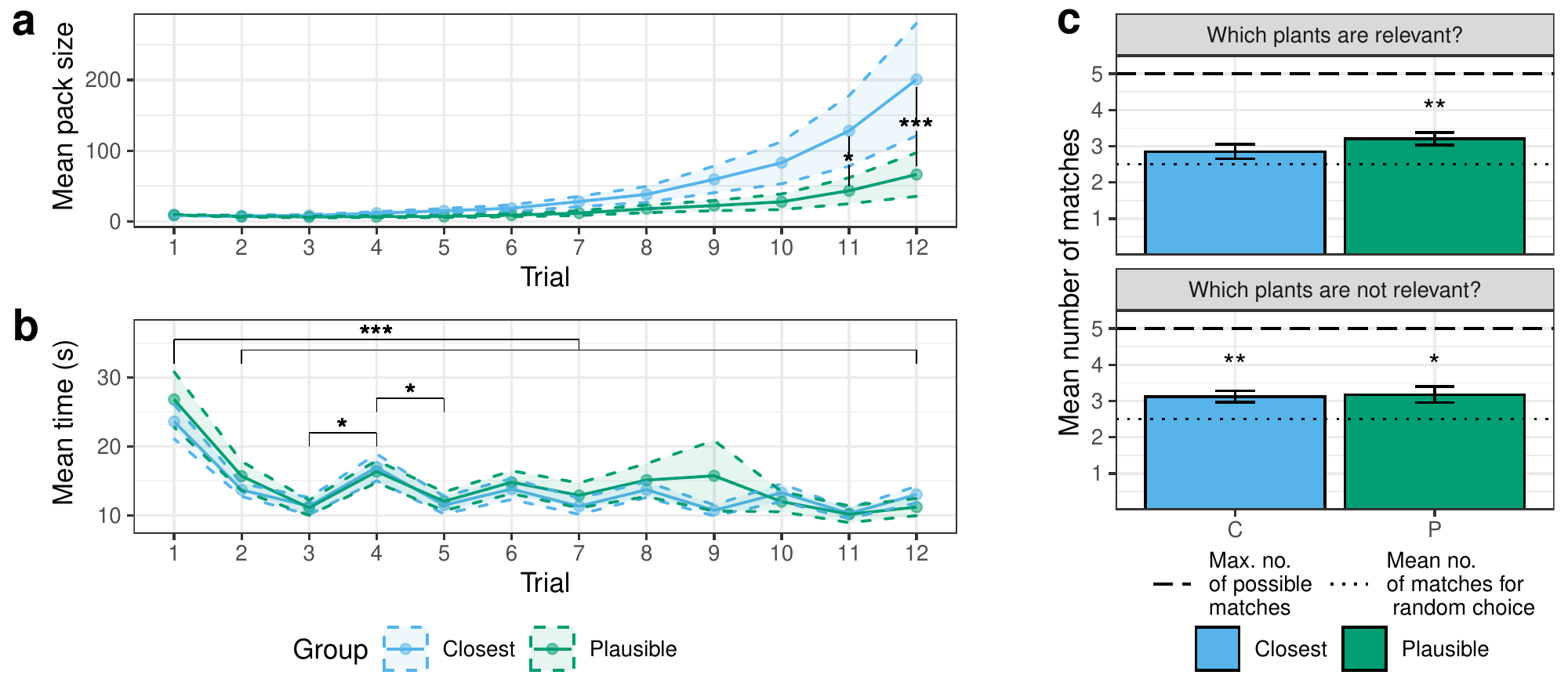}
   \caption{Development of (\textbf{a}) mean pack size per group by trial, (\textbf{b}) mean decision time per group by trial, and (\textbf{c}) mean number of matches between user judgments and ground truth for survey items assessing relevant plants and irrelevant plants, respectively. Shaded areas in (\textbf{a}), (\textbf{b}), and error bars in (\textbf{c}) denote the standard error of the mean. Asterisks denote statistical significance (\textit{p} < .05 (*), \textit{p} < .01 (**), and \textit{p} < .001 (***), respectively. Asterisks in (\textbf{c}) denote statistical significance from expected value for random behavior.}
   \label{fig:hyp1}
 \end{figure}

Hypothesis 1 postulates that users in the \textit{plausible} condition outperform users in the \textit{closest} condition.
To statistically assess this hypothesis, we compare data from participants in both groups in terms of pack size produced over time, decision time, and matches between ground truth and indicated plants.
Figure \ref{fig:hyp1}a shows the development of average pack size as well as average decision time per group. 
Strikingly, the data suggests that participants in the \textit{closest}, not the \textit{plausible}, condition performed better. 
This effect is confirmed by the significant interaction of factors \textit{trial number} and \textit{group} (\textit{F}(11,792) = 2.119, \textit{p} = .017, $\eta_{\text{p}}^{2}$ = 0.029) in the corresponding linear mixed effects model. The follow-up analysis reveals significant differences between groups in trial 11 (\textit{t}(472) = 4.040, \textit{p} = .012, \textit{d} = 0.693) and trial 12 (\textit{t}(472) = 2.530, \textit{p} \textless .001, \textit{d} = 1.101).
Additionally, there is a highly significant main effect of trial number (\textit{F}(11,792) = 7.585, \textit{p} \textless .001, $\eta_{\text{p}}^{2}$ = 0.095), but no significant main effect of group (\textit{F}(11,72) = 2.586, \textit{p} = .112, $\eta_{\text{p}}^{2}$ = 0.035).

Participants in both groups showed a marked decrease in decision time over the curse of the study, already apparent after the first trial (Figure \ref{fig:hyp1}b).
The significant main effect of factor \textit{trial number} (\textit{F}(11,792) = 14.818, \textit{p} \textless .001, $\eta_{\text{p}}^{2}$ = 0.171) confirms this observation.
Corresponding post-hoc analyses show significant differences between trial 1 and all other trials (all \textit{t}(792) \textgreater 5.900, \textit{p} \textless .001, \textit{d} \textgreater 1.200), between trial 3 and 4 (\textit{t}(792) = 3.765, \textit{p} = .012, \textit{d} = 0.621), and between trials 4 and 5 (\textit{t}(792) = 3.395, \textit{p} = .048, \textit{d} = 0.560).
Neither the main effect of factor \textit{group} (\textit{F}(11,72) = 0.235, \textit{p} = .630, $\eta_{\text{p}}^{2}$ = 0.003), nor the interaction between factors \textit{trial number} and \textit{group} (\textit{F}(11,792) = 0.897, \textit{p} = .543, $\eta_{\text{p}}^{2}$ = 0.012) reach significance.

In terms of mean number of matches between user judgments of plant relevance for task success and the ground truth, users in both groups performed comparably both for relevant 
(\textit{closest}: mean number of matches = 2.850 $\pm$ 0.198 \textit{SE}; \textit{plausible}: mean number of matches = 3.206 $\pm$ 0.178 \textit{SE}; \textit{U} = 781, \textit{p} = .255, \textit{r} = .054)
and irrelevant plants (closest: mean number of matches = 3.125 $\pm$ 0.157 \textit{SE}), plausible: mean number of matches = 3.177 $\pm$ 0.217 \textit{SE}; \textit{U} = 721.5, \textit{p} = .643, \textit{r} = .054).
While groups do not differ in terms of matches between user judgments of plant relevance, we find significant differences between mean group responses compared to the expected value given random responses (\ie, expected mean number of matches = 2.500): 
Users in the \textit{computationally plausible} group have significantly more matches than random for both items (relevant plants: \textit{W} = 481, \textit{p} = .005, \textit{r} = .550; irrelevant plants: \textit{W} = 459.5, \textit{p} = .020, \textit{r} = .483).
Users in the \textit{closest} group have significantly more matches than random when identifying irrelevant (\textit{W} = 659.5, \textit{p} = .002, \textit{r} = .544), but not relevant (\textit{W} = 536.5, \textit{p} = .331, \textit{r} = .274) plants.

Thus, we cannot verify our hypothesis that \textit{computationally plausible \glspl{CFE}} facilitate learning. On the contrary, the development of pack size between the groups points to the opposite effect of \textit{closest \glspl{CFE}} being more beneficial for users than \textit{computationally plausible} ones.

\subsection{Do Computationally Plausible CFEs Increase User's Subjective Understanding?}
\begin{figure}[t]
   \centering
   \includegraphics[width=\textwidth]{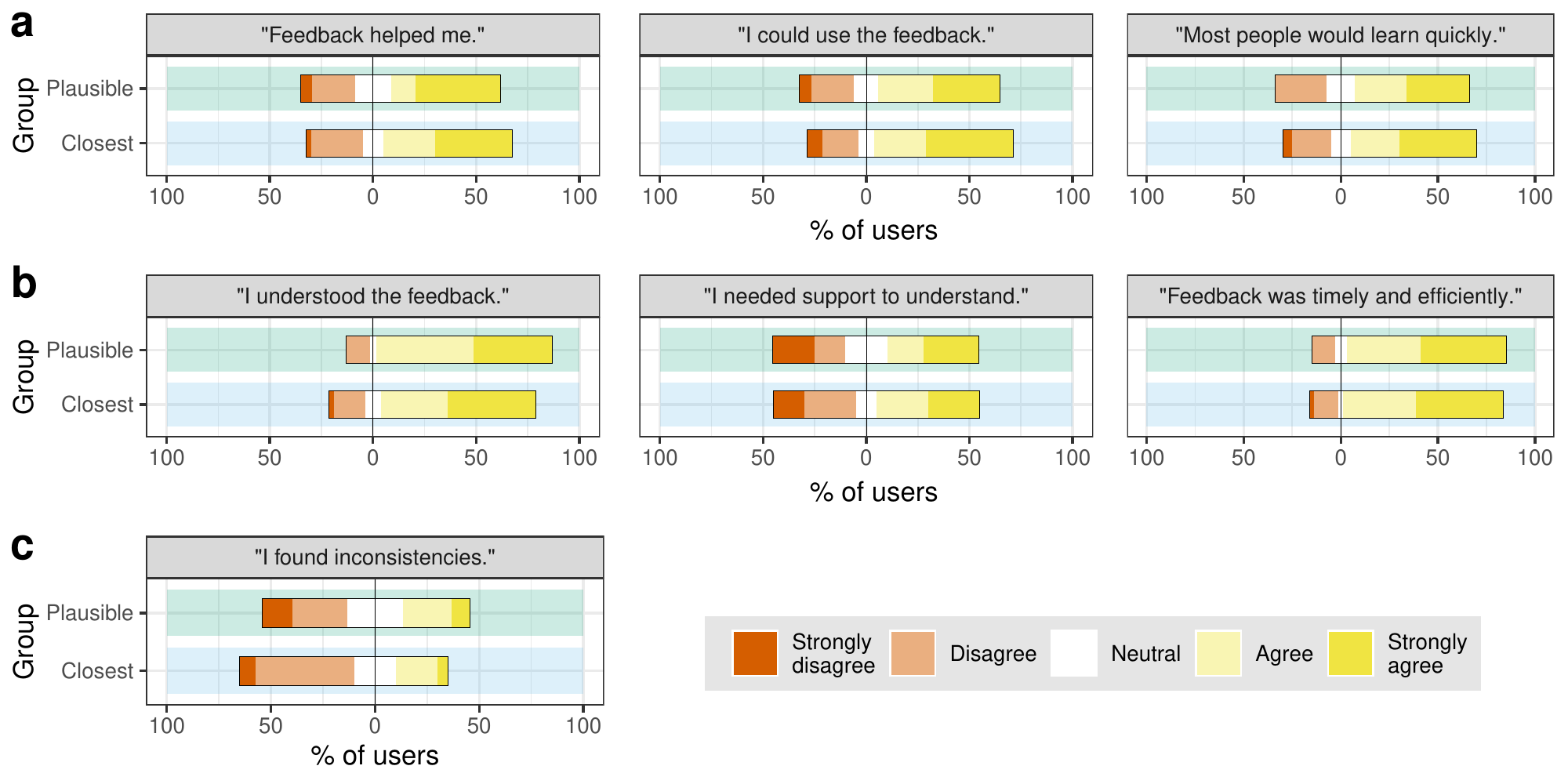}
   \caption{Overview of user judgments in post-game survey per group, adapted from ~\citep{holzinger_measuring_2020}. (\textbf{a}) depicts user replies in survey items relevant for hypothesis 2, (\textbf{b}) depicts user replies in survey items relevant for hypothesis 3, and (\textbf{c}) depcits replies relevant for our last exploratory analysis. Distributions did not differ significantly between groups for any of the items (all \textit{p} \textgreater .05).}
   \label{fig:survey}
 \end{figure}

To assess hypothesis 2, we analyze participant judgments on relevant survey items.
Visual assessment suggests that there is little variation in terms of user responses between groups (Figure \ref{fig:survey}a), confirmed by our statistical assessment. 
Groups do not statistically differ when judging whether presented \gls{CFE} feedback was helpful to increase pack size (\textit{closest} condition: \textit{M} = 3.700 $\pm$ 1.285 \textit{SE}; \textit{plausible} condition: \textit{M} = 3.636 $\pm$ 0.242 \textit{SE}; \textit{U} = 656, \textit{p} = .968, \textit{r} = .005).
Likewise, we do not detect significant group differences in terms of subjective usability (\textit{closest} condition: \textit{M} = 3.775 $\pm$ 0.216 \textit{SE}; \textit{plausible} condition: \textit{M} = 3.606 $\pm$ 0.230 \textit{SE}; \textit{U} = 603, \textit{p} = .513, \textit{r} = .077).
In addition, there is no significant difference between groups for estimated usefulness of explanations for others (\textit{closest} condition: \textit{M} = 3.750 $\pm$ 0.208 \textit{SE}; \textit{plausible} condition: \textit{M} = 3.647 $\pm$ 0.206 \textit{SE}; \textit{U} = 637, \textit{p} = .631, \textit{r} = .056).

\subsection{Does Mode of Presentation have an Impact?}
As postulated in hypothesis 3, we do not observe group differences between conditions in terms of understanding the explanations as such (Figure \ref{fig:survey}b). 
A considerable proportion of both groups responds positively about understanding the feedback, not differing significantly in their responses (\textit{closest} condition: \textit{M} = 3.975 $\pm$ 0.184 \textit{SE}; \textit{plausible} condition: \textit{M} = 4.118 $\pm$ 0.162 \textit{SE}; \textit{U} = 773.5, \textit{p} = .200 \textit{r} = .149).
In terms of needing support for understanding, both groups reply with a similar response pattern (\textit{closest} condition: \textit{M} = 3.200 $\pm$ 0.230 \textit{SE}; \textit{plausible} condition: \textit{M} = 3.147 $\pm$ 0.257 \textit{SE}; \textit{U} = 667, \textit{p} = .890 \textit{r} = .016).
Similarly, user judgments on timing and efficacy of  CFE presentation are consistently high across groups (\textit{closest} condition: \textit{M} = 4.100$\pm$0.175 \textit{SE}; \textit{plausible} condition: \textit{M} = 4.147$\pm$0.170 \textit{SE}; \textit{U} = 680.5, \textit{p} = 1 \textit{r} = .000).

\subsection{Exploratory Analysis}
More than half of all users in both groups do not report to having detected any inconsistencies in the \glspl{CFE} provided (\textit{closest} condition: \textit{M} = 2.675 $\pm$ 0.166 \textit{SE}; \textit{plausible} condition: \textit{M} = 2.853 $\pm$ 0.207 \textit{SE}; \textit{U} = 743, \textit{p} = .480, \textit{r} = .082).

\section{Discussion}\label{sec:discussion}

In this work, we investigate effects of implementing a plausibility constraint on computed \glspl{CFE} for \gls{ML} models on user performance in an iterative learning task in an abstract domain. 
The employed constraint limits the set of possible solutions to the training data.
We measure understanding and usability of explanations in terms of two objective behavioral variables, \ie, task performance and decision time, and several subjective self-reports. 
Our results reveal a range of valuable insights with important implications for \gls{XAI} in application.

%%%% Hypothesis 1 - Do plausible CFEs facilitate learning?
First, we cannot verify our initial supposition that \textit{computationally plausible} \glspl{CFE} facilitate learning in the current setting. 
Intriguingly, we observe the opposite effect: users in the \textit{closest} condition generated larger pack sizes than users in the \textit{computationally plausible} condition.
A likely reason for this observation may be that the current study revolves around an abstract scenario of feeding aliens. 
Given psychological interpretations of plausibility as probability~\citep{pezdek_is_2006, de_brigard_remembering_2013}, simply restricting \glspl{CFE} to items from the training set cannot help participants that lack an informed mental representation of the current state of this alien world. 
For a novice user at the onset of the study, any counterfactual is equally likely.

% Why is closer better - maybe higher comparative similarity!
We may turn to the definition of psychological plausibility as comparative similarity~\citep{lewis_counterfactuals_1973, stanley_counterfactual_2017} for a possible explanation why users in the \textit{closest} group showed significantly superior performance. 
Classically, \glspl{CFE} are penalized if they deviate from the requested prediction~\citep{wachter_counterfactual_2017}, resulting in \textit{closest \glspl{CFE}} that differ minimally from the user's input.
This concept resembles the view that psychologically plausible counterfactuals come from worlds that are minimally different from reality~\citep{lewis_counterfactuals_1973}.
Empirical evidence highlighting the close relation between perceived plausibility and perceived similarity between counterfactual and factual state supports this notion~\citep{stanley_counterfactual_2017, de_brigard_perceived_2021}.
In contrast, the computational plausibility constraint rejects \glspl{CFE} that are not part of the training set, even if they are minimal.
Consequently, users in the \textit{plausible} condition encounter larger differences between provided explanations and their input, at odds with the idea of plausibility as comparative similarity.
Conversely, users in the closest condition might have experienced their \glspl{CFE} as more psychologically plausible than the computationally plausible version.
Further, upon seeing \textit{closest \glspl{CFE}}, users might get a feeling of ``just missing'', inducing negative affect that strongly motivates improvement~\citep{medvec_when_1997, markman_reflection_2003}.
Additionally, the larger discrepancies between factual and counterfactual state in the \textit{plausible} condition might increase the mental load on users, potentially hampering learning.
Future studies need to disentangle contributions of these factors.

% Refer to Lim, 2009 - they did not find any effects for What-if explanations, 
Intriguingly, our results are at odds with empirical findings indicating that \glspl{CFE} for intelligent systems do not improve user's task performance~\citep{lim_why_2009, van_der_waa_evaluating_2021}.
~\citeauthor{lim_why_2009} assessed the effectiveness of different explanation modes for context-aware systems~\citep{lim_why_2009}. 
In their study, performance of users receiving counterfactual style \textit{what-if}-explanations was indistinguishable from that of users getting no explanations what-so-ever.
In contrast, users in our study indeed show learning after receiving \glspl{CFE}.
Interestingly, their task resembles ours in so far that they also employed an abstract domain: users chose values of non-specific features (labelled \textit{A}, \textit{B} and \textit{C}), relating to a non-specific prediction (\textit{a} or \textit{b}).
However, while also dealing with an abstract task (\ie, feeding aliens), our users have a tangible goal (\ie, make the pack grow).
Further, we refrain from separating learning and testing as in ~\citeauthor{lim_why_2009}, where users went through an initial evaluation section receiving explanation after explanation.
Our design is far more interactive, with different rounds of user action and feedback.
This is in line with evidence from educational science, suggesting that learner's level of engagement relates to learning outcome, with interactive activities granting deepest understanding~\citep{chi_icap_2014}.
%These vital differences may make our design more effective.
Thus, including goal-directed and interactive settings may potentially be vital facets of effective usability studies. 
We suggest that future research designs need to pay special attention to these aspects in order to accurately evaluate \gls{XAI} approaches.

% We did not find timing differences
Beyond task performance, we quantify learning success in terms of user's decision time and their ability to explicitly state which plants were relevant. % (survey items 1 and 2).
Both measures do not reveal significant group differences.
In terms of user's decision time, both groups show significant speed-up already after Trial 1 (Figure \ref{fig:hyp1}a).
This initial time decrease likely reflects how participants learn to work with the game interface efficiently.
Increased reaction time as a marker of learning is a classical insight from experimental psychology~\citep{logan_shapes_1992}, indicating that both groups did indeed learn in the current setting. 
It is possible that the complex task we devised with its elaborate game-like setting was not sensitive enough or too short to pick up in subtle group differences usually linked to more simple, extensive reaction time experiments.

% We did not find differences in relevance judgments
Users in the closest group show superior performance, however, they are not able to state more explicitly which plants were relevant or irrelevant for the given task. 
With this, our study replicates a recent observation that objective measures (\ie, task performance) do not necessarily correlate with self-reports reflecting system understanding~\citep{van_der_waa_evaluating_2021}.
Participants in both groups made three out of five correct choices on average (Figure \ref{fig:hyp1}c), in part significantly exceeding the number expected in case of random behavior.
Thus, both groups showed some--yet imperfect--explicit understanding of the underlying system.
Potentially, users may rely on their initial mental model of the appropriate alien diet, allowing them to make advantageous feeding choices relatively quickly. However, in this initial stage, it may still be insufficient to allow clear and explicit differentiation between relevant and irrelevant features at the end of the study.

%%%% Hypothesis 2 - Do plausible CFEs increase user's subjective understanding?
Besides effects on task performance, we do not detect any statistically meaningful differences between the two groups under investigation, predominantly affecting the evaluation of user judgments in the survey. 
It is clear that these observations have to be taken with the care generally devoted to null effects, calling for cautious interpretation. 
Still, regarding the general trends for individual survey items is informative.
We cannot verify our second hypothesis, as users did not differ depending on group in terms of subjective helpfulness and usability. 
Still, we note that the majority of users in both groups respond with agreement or strong agreement in the respective items (Figure \ref{fig:survey}a).
This supports the notion that \glspl{CFE} are indeed subjectively intuitive and usable for lay users, also when used in an abstract setting.

%%%% Hypothesis 3 - Does mode of presentation have an impact?
A major challenge for effective user designs comparing different approaches is keeping conditions highly comparable, with the sole exception of the experimental manipulation.
User judgments of general understandability and presentation mode of \glspl{CFE} inspire confidence that we achieved this level of control with our Alien Zoo design. 
In fact, the respective items elicit the highest user judgments out of all survey responses, with agreement values close to 90\%.% (questionnaire items 3, 4 + 10). So this is also control to make sure groups don’t differ in a weird way.
~High agreement across both groups leads us to conclude that mode of \gls{CFE} presentation does not have an impact when comparing users experience \textit{closest} vs. \textit{plausible} \glspl{CFE}, validating our hypothesis 3.
Thus, the Alien Zoo design does not just elucidate benefits of providing \glspl{CFE} compared to no explanation~\citep{kuhl_lets_2022}, it is also suitable for user evaluations of \gls{CFE} methods, a yet vastly understudied aspect in the field of \gls{XAI}~\citep{keane_if_2021}.

%%%% Exploratory Analysis
Finally, the majority of users indicate that they fail to find inconsistencies in the \glspl{CFE} provided.
Thus, we can rule out that our code generated irregular or even contradictory explanations, confounding the observed group differences.

\subsection{Limitations \& Future Work}\label{subsec:limitations-future-work}

Several limitations warrant caution when interpreting our results beyond the scope of this work.
%What are limitations?
%% does it generalize? NOT beyond the setting (nocive users with an abstract domain, trying to learn something)
Critical design choices in any \gls{XAI} evaluation include the reason for explaining, and the target group~\cite{adadi_peeking_2018}.
The current results may only be generalized to cases with the same motivation for explaining (\ie, to `explore') as well as the intended audience (\ie, novice users).
Other motives and applications addressing more specific target groups call for independent studies.

%% sample size: we pre-computed it, but had to throw out a portion - does it still hold up?
We excluded data from 26 out of 100 participants to meet a priori quality criteria.
Such participant attrition common, especially in web-based studies. 
As smaller sample sizes always mean a loss of statistical power, we factored in this issue in an a priori power analysis. 
Yet, the effect size of the significant interaction between factors \textit{trial number} and \textit{group} remains relatively small. 
Hence, the results from this work await confirmation in larger follow-up studies.
 
% was our survey suitable? 
None of the survey items revealed significant group effects, in line with a previous account of diverging trends between objective measures and self-reports~\citep{van_der_waa_evaluating_2021}. 
This may reflect a more general tendency in human evaluation of system understanding.
Alternatively, however, we may call into question the efficacy of instruments applied to assess user experience. 
To date, there is no standard inventory for assessing subjective usability in \gls{XAI} research.
We adapted the System Causability Scale \cite{holzinger_measuring_2020} to determine subjective usability of presented \glspl{CFE}.
Yet, there is no large scale validation of this measure.
One potential shortcoming may be a lack of sensitivity to subtle group differences.

%% difficult to durectly translate for real world applications
Moreover, the current scenario in its present condition may be difficult to translate to specific real-world applications.
The lack of realism offers full algorithmic recourse~\citep{karimi_survey_2020}: all changes are feasible (\ie, doable for the participant), and all changes in features are independent (\ie, a user can change plant 1, and this will have no long-time effect on plant 2).
In real life scenarios, this is barely the case (\eg, a bank customer might never be able to get younger to get a loan; yearly income also affects savings). 
Thus, our example is much more artificial, and we suggest applying iterative learning designs in more realistic, real-world scenarios as an exciting avenue for future work.

%% What did ppl do that did not show high levels of learning?
Users in our study play an elaborate online game, with a detailed user interface, and several consecutive scenes.
Designed to be maximally engaging as to ensure participant compliance, the amount of information displayed may overwhelm some participants.
This could explain inferior performance of a small proportion of participants, like those who disagree with the notion that feedback presentation was timely and efficiently.

% FUTURE WORK
% important ToDo: Comparisons with no-explanations + quatschexplanations
Recent evidence suggests an added benefit of providing users with \glspl{CFE} over no explanations to understand the behavior of an unknown system~\citep{van_der_waa_evaluating_2021}.
The current work expands this insight by a direct comparison of two different approaches for \gls{CFE} computation. 
While our results suggests the suitability of our Alien Zoo design, further validation studies must delineate potential shortcomings.
For instance, a crucial validation step of the design itself concerns comparisons of valid \glspl{CFE} with no explanations or non-sensical ones.

Beyond such a fundamental investigation, the Alien Zoo design lends itself to be easily modified.
Possible adaptations may incorporate data with different dynamics, use other \gls{ML} models, or compare other \gls{CFE} approaches.
Thus, the design has tremendous potential to answer open questions in the domain of \gls{XAI}.
For instance, future work may explore the impact of distinct psychometric properties on performance. 
A small-scale user study
%based on data from 38 participants 
suggests an effect of individual personality traits on user's ability to make sense of an \gls{ML} system's output, and understanding the generation process, respectively~\citep{gleaves_role_2020}.
It remains to be shown how personal attributes relate to usability judgments of \glspl{CFE}.

Further, it is conceivable that users may prefer to receive explanations on demand, rather than continuously at prescribed intervals.
There is abundant room for further progress in determining whether explicitly requesting \glspl{CFE} may improve task performance, and how users would make use of their control over the explanation intervals.

Finally, we successfully demonstrate usefulness of \glspl{CFE} for the current task, indicating a certain degree of intuitiveness or plausibility connected to them. 
Future investigations may tackle whether \glspl{CFE} cause users to fall prey to a plausibility fallacy, coming to trust biased or unfair ML models just because they are coupled with intuitive explanations.

\subsection{Conclusions}\label{subsec:conclusion}
In this work, we present a controlled study comparing user performance and usability judgments of \glspl{CFE} in an iterative learning design.
We focus on potential group effects driven by receiving either \textit{closest \glspl{CFE}} that are minimally different from the user's input, compared to \textit{computationally plausible} ones, limited to instances found in the training data.
We develop an accessible game-like experimental design revolving around an abstract scenario, suitable for novice users.
Our design demonstrates learning in both groups, highlighting the power of interactive and goal-directed tasks for user evaluations of \gls{CFE} methods, a yet vastly understudied aspect in the field of \gls{XAI}.
Moreover, our findings suggest that novice users benefit more from receiving \textit{closest} than \textit{computationally plausible \glspl{CFE}}.
This supports the view of plausibility as comparative similarity rather than probability in cases where users lack an accurate mental model to build on.
In sum, our work emphasizes yet again that theoretical approaches proposing explanation techniques for \gls{ML} models and user-based validations thereof need to go hand in hand.
Researchers designing \gls{XAI} approaches need to bear in mind human behavior, preferences and mental models, to build on a solid foundation to effectively benefit the end user.

% Acknowledgements:
\section{Acknowledgements}
The authors would like to thank Johannes Kummert for support with server set up and maintenance.

\textbf{Funding/Support:} This research was supported by the \gls{BMBF}~~and the \gls{DLR}~~through grant agreement no.:~01IS18041 A (project ITS.ML), and the VW-Foundation~~(project IMPACT).

%%%%
% References
%%%%

\AtNextBibliography{\raggedright\small}
\printbibliography

\end{document}